\documentclass[10pt,twocolumn,letterpaper]{article}

\usepackage{iccv}
\usepackage{times}
\usepackage{epsfig}
\usepackage{graphicx}
\usepackage{amsmath}
\usepackage{amssymb}
\usepackage{subfigure}
\usepackage{multicol}
\usepackage{multirow}

\usepackage[pagebackref=true,breaklinks=true,letterpaper=true,colorlinks,bookmarks=false]{hyperref}

\iccvfinalcopy 

\def\iccvPaperID{24} 

\ificcvfinal\fi
\begin{document}

\title{A Realistic Face-to-Face Conversation System based on Deep Neural Networks}
	
\author{Zezhou Chen\\
		cloudminds\\
		{\tt\small chenzezhou007@aliyun.com}
		\and
		Zhaoxiang Liu\thanks{corresponding author}\\
		cloudminds\\
		{\tt\small robin.liu@cloudminds.com}
	    \and
		Huan Hu\\
		cloudminds\\
		{\tt\small hans.hu@cloudminds.com}
	    \and
		Jinqiang Bai\\
		Beihang University\\
		{\tt\small baijinqiang@buaa.edu.cn}
        \and
		Shiguo Lian\\
		cloudminds\\
		{\tt\small  sg\_lian@163.com}
	    \and
		Fuyuan Shi\\
		cloudminds\\
		{\tt\small fuyuan.shi@cloudminds.com}
	    \and
		Kai Wang\\
		cloudminds\\
		{\tt\small kai.wang@cloudminds.com}
}
	
\maketitle
\begin{abstract}
	To improve the experiences of face-to-face conversation with avatar, this paper presents a novel conversation system. It is composed of two sequence-to-sequence models respectively for listening and speaking and a Generative Adversarial Network (GAN) based realistic avatar synthesizer. The models exploit the facial action and head pose to learn natural human reactions. Based on the models' output, the synthesizer uses the Pixel2Pixel model to generate realistic facial images. To show the improvement of our system, we use a 3D model based avatar driving scheme as a reference. We train and evaluate our neural networks with the data from ESPN shows. Experimental results show that our conversation system can generate natural facial reactions and realistic facial images.
\end{abstract}

\section{Introduction}

Recently, virtual assistants have been playing a more and more important role in our daily life, such as the question-answering assistant Apple Siri and Amazon Alexa, which may be the most popular virtual assistants at now. However, they provide only verbal response, lacking of nonverbal feedback.

Alternatively, some 2D or 3D avatars have been introduced to virtual assistants \cite{23,47,Chu2018AFN}, which can provide both verbal and nonverbal interaction experiences. Nevertheless, few of them can exhibit realistic nonverbal reactions while speaking, \eg, realistic facial expressions. What is more, no natural reactions are exhibited during the listening phase when the avatars are talking with human\cite{Chu2018AFN}. Generally, they either keep still or randomly react within a series of predefined actions while listening to human. That is far from human's expectation on realistic and nature conversation.

\begin{figure}
	\begin{center}
		\includegraphics[width=0.9\linewidth]{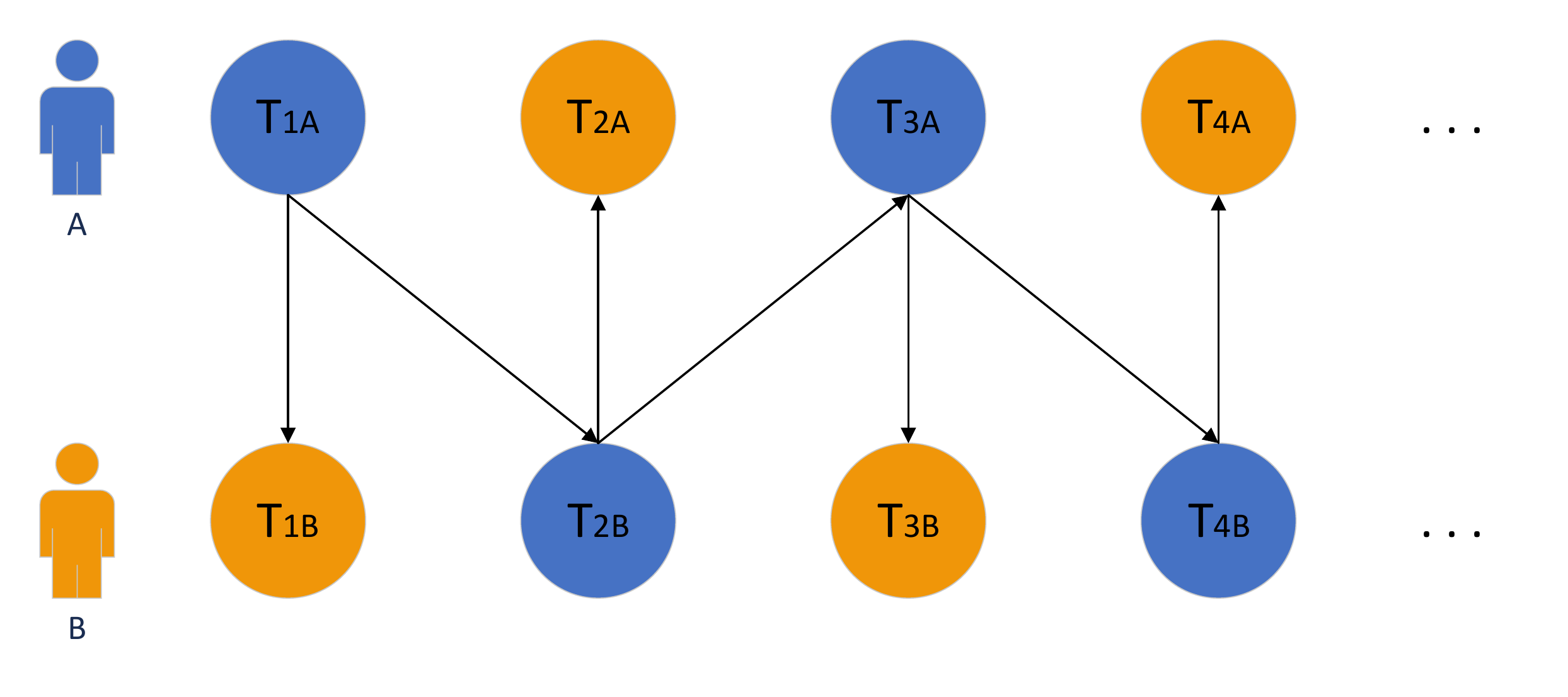}
	\end{center}
	\caption{Regularity of human-human conversation. $T_{iA}$ and $T_{iB}$ $(i=1,2,3,...)$ denotes person A and B is speaking or listening in the $i$-th time period respectively. Blue blocks indicate that the person is a speaker, while orange blocks a listener. Arrows imply the dependence between different roles in different time periods.}
	\label{fig:1}
\end{figure}

We aim to fill the gap between present virtual assistants and the demand of realistic and natural interaction. Our work is based on the observations of conversational regularity illustrated in Figure \ref{fig:1}. In human-human communications, the two roles, listener and speaker, are alternative between the two parties involved in the conversation. The speaker often makes verbal actions (speech) and nonverbal actions (head movements and facial expressions) simultaneously. While  listening to the speaker, the listener receives verbal information as well as nonverbal cues and gives nonverbal feedback. Their roles exchange when the previous speaker stops talking. And the new speaker makes verbal and nonverbal responses accordingly. This procedure carries on repeatedly until the conversation ends.

Motivated by the great success of Sequence-to-Sequence (Seq2Seq) network \cite{25} and Generative Adversarial Network(GAN) \cite{12,13,14,15,16,17}, we propose a novel face-to-face conversation system which consists of seq2seq based listening and speaking models and a GAN based realistic avatar synthesizer. The listening and speaking models serve as generators of natural facial reactions and head pose. The realistic avatar synthesizer is used to produce a realistic portrait looks like a real human. Generally, the facial action and head pose to some extent imply the intentions and emotions of both speaker and listener. The listening model takes speaker's facial action and head pose as input and generates facial action and head pose as listener's nonverbal response. The speaking model takes the verbal response as input and generates facial action and head pose as speaker's nonverbal accompaniment. The realistic avatar synthesizer takes the outputs of alternative listening model and speaking model as input and produces a realistic image sequence for avatar. Compared to traditional 3D data based realistic face synthesizer, our avatar synthesizer is free from the need for expensive 3D or motion capture data.

The rest of the paper is organized as follows. Some related works are reviewed in Section \ref{Review}. In Section \ref{System}, the proposed conversation system is presented in detail. Experimental results and discussions are given in Section \ref{results}. Finally, Section \ref{summary} draws the conclusion.

\section{Related works}\label{Review}

Generally, a realistic face-to-face conversation system is able to understand the human's speech and action, then decide how to response in verbal and nonverbal manners, and finally drive the avatar. Although few mature end-to-end systems exist till now, there have been many works related to each of the steps.

To understand humans in interaction, some means have been proposed, including head pose estimation \cite{1,3,4,DBLP:journals/corr/abs-1904.13102}, gesture recognition \cite{6}, gaze tracking \cite{7}, and multi-modal fusion \cite{2}. Especially, analyzing facial actions to get emotion information is important for face-to-face conversation. Generally, face landmarks \cite{44} are used for facial analysis, which tells the 2D or 3D positions of key points of the face. Alternatively, action unit has been widely used for face expression analysis or face avatar synthesis \cite{8,9,10,11,35}, which represents face components' local movements. Different from face landmark, action unit contains only the relative motion of face components while no head pose or face size. Additionally, action unit, as a nonverbal feature, can also be used for verbal analysis, \eg, lip reading \cite{8}. It means that action unit contains both verbal and nonverbal information. Motivated by it, in this paper, we extract action unit and head pose from a face image as face features.

For verbal chatting, various conversation models have been widely used. For example, \cite{26} proposed a translation model based on the encoder-decoder architecture. \cite{25} presented a LSTM \cite{22} based seq2seq structure to solve the training problem of long sequences. However, for nonverbal conversation or verbal-nonverbal combined conversion, there is few mature works. The latest one \cite{26} adapted the seq2seq structure from verbal domain to nonverbal domain, and proved the seq2seq model works for nonverbal scenarios, although it considered only the speaking phase. Inspired by the work, we also use the seq2seq structure to design nonverbal models, including both the speaking model and listening model, which produces facial action units and head pose for avatar.

For avatar synthesis, the action unit based avatar driving technique \cite{6,Chu2018AFN} has often been used, which activates the predefined 2D or 3D avatar by an action unit sequence. For example, the work in \cite{Chu2018AFN} also proposed a face-to-face conversation model which could drive a 3D avatar to answer human's questions. However, avatars are either not realistic enough or difficult to obtain by such means as accurate 3D model scanning. What is more, no natural reactions are exhibited for the 3D avatar during the listening phase in \cite{Chu2018AFN}. Fortunately, due to the rapid development of Generative Adversarial Networks (GAN) in image generation \cite{12}, it is able to synthesize photo-realistic facial image without complex 3D models. For example, Pix2PixHD \cite{16} and CycleGAN \cite{14} offer appealing and succinct alternative for facial image synthesis. Recently, Chan \etal applied Pix2PixHD to learn a motion transformation between two video subjects \cite{15}. Wang \etal presented the framework of Vid2Vid \cite{17} to learn mappings between different videos. Motivated by these works \cite{15,16,17,Chu2018AFN,DBLP:journals/corr/abs-1908.07262,DBLP:journals/corr/abs-1908-06607} , we also adopt the GAN to generate the realistic avatar with facial action units and head pose as input.

\begin{figure*}[htbp]
	\begin{center}
		\includegraphics[scale=0.4]{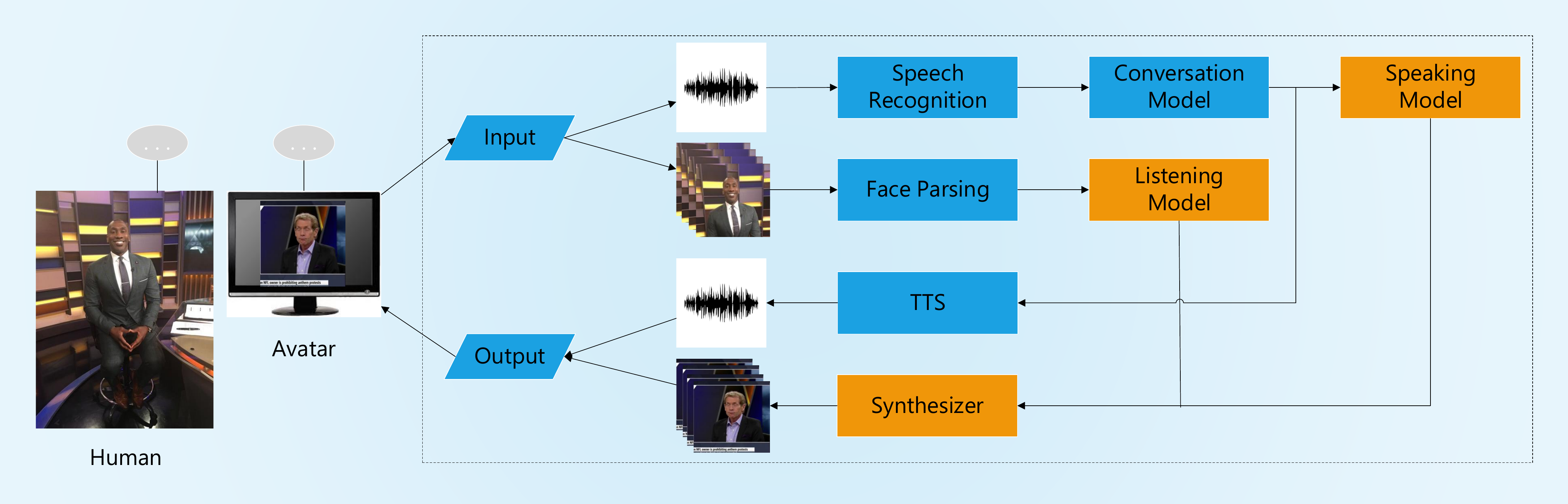}
	\end{center}
	\caption{Overview of the proposed realistic face-to-face conversation system. Blocks colored orange indicate our contributions.}
	\label{fig:2}
\end{figure*}

\section{The proposed system}\label{System}
\subsection{System overview}
In this paper, we propose a novel face-to-face conversation system. As shown in Figure \ref{fig:2}, there are two phases in our system: one is for listening, and the other for speaking. When the virtual avatar is communicating with human, the speech audio and face images are input into our system. The face images are processed by face parsing module that produces face information including facial action units and face pose. This information is then passed into the Seq2Seq based listening model whose output is fed into the avatar synthesizer to produce realistic face images as nonverbal reactions when the virtual avatar is listening. The speech audio is transformed into text using the speech recognition algorithm proposed in \cite{40,41}, and then passed into the conversation module for generating response sentence. The response sentence is passed to the Text-to-Speech (TTS) module proposed in \cite{42,43} to produce synthesized speech. At the same time, the response sentence is passed to the Seq2Seq based speaking model whose output is also input into the avatar synthesizer to produce realistic face images as the accompaniment of speech for the speaking phase. The avatar synthesizer serves listening and speaking alternatively during the whole conversation.

To train our system, we collect 700 videos of ESPN shows (First Take and Undisputed) from youtube that contain the scenes of face-to-face conversation.

\subsection{Face parsing}
To get the facial feature representation, we have several choices such as facial landmark (FL) \cite{20}, action unit \cite{2,5} and head-pose. FL carries detailed information on facial features and outlines. However, it has three disadvantages. First, FL lacks information on some areas such as forehead and cheeks. Second, FL cannot directly reflect the movement of facial muscles. Third, compared to the action unit, FL is more personalized and we want our high-level features to carry more generalized information. So we combine action-unit and head-pose (AU+POSE) to represent facial features. Compared with the combination of head-pose and text or the combination of text and action-unit, AU+POSE retains sufficient semantic information. In the listening phase, AU+POSE can maintain the smooth of dataflow in the deep network because no extra coding is needed, and this is conducive to the training of the entire system.

The Open-Face \cite{19} can consistently predict the action units using Support Vector Machine (SVM) and Support Vector Regression (SVR) \cite{39} with concatenated HoG feature \cite{37} and facial shape, and accurately estimate head pose by solving PnP problem. Thus, we use it to extract AU+POSE from both the speaker and listener. The AU+POSE is a 20-D vector which comprises of a 17-D AU and a 3-D head pose. On the acquisition of text data, the sentence is obtained directly from the video with python library AutoSub and TTS.

\subsection{Listening model}
The listening model takes the AU+POSE of speaker as input and outputs the AU+POSE of listener. We base our listening model on the Seq2Seq \cite{25} architecture (as shown in Figure \ref{fig:3}).

\textbf{Listening encoder:}
The AU+POSE sequence extracted from the face sequence of speaker is employed as the input of the listening encoder.
Let $d_1,d_2,\ldots,d_n$ represent the AU+POSE sequence extracted from the consecutive $n$ face images, and $h_1,h_2,\ldots,h_n$ represent the hidden transitions of encoder for each input frame. The model is defined as

\begin{equation}\label{eqn:1}
\begin{aligned}
\begin{array}{l}
i^t = \sigma \left(W_i\left[h^{t-1},d^t\right] + b_i\right)\\
f^t = \sigma \left(W_f\left[h^{t-1},d^t\right] + b_f\right)\\
o^t = \sigma \left(W_o\left[h^{t-1},d^t\right] + b_o\right)\\
c^{-t} =relu6\left(W_c\left[h^{t-1},d^t\right] + b_c\right)\\
c^t = f_t\odot c_{t-1} + i_t \odot c^{-t}\\
h_{enc}^t = o_t \odot relu6(c_t)
\end{array}
\end{aligned}
\end{equation}where $\sigma$ denotes the sigmoid function, $relu6$ is the piece-wise linear function, $\odot$ denotes the element-wise multiplication, $i_t,f_t,o_t$ represent input gate, forget gate, output gate of $t$-th LSTM unit respectively. $c_t$ and $h_t$ are the $t$-th cell state and hidden state. $W$ and $b$ are the trainable weights.

\textbf{Listening decoder:}
The decoder model in the listening phase maintains the same $n$ serialized outputs as the inputs. The output can be formulated as

\begin{equation}\label{eqn:2}
\begin{aligned}
h_{dec}^l = LSTM\left(h_{dec}^{l-1}|h_{enc},y^{l-1}\right)\\
\end{aligned}
\end{equation}
where $h_{dec}$ is the hidden state of $l$-th decoder, and $y^{l-1}$ is the output AU+PUSE of the $(l-1)$-th decoder.

\subsection{Speaking model}
\begin{figure*}[htbp]
	\begin{center}
		\includegraphics[scale=0.5]{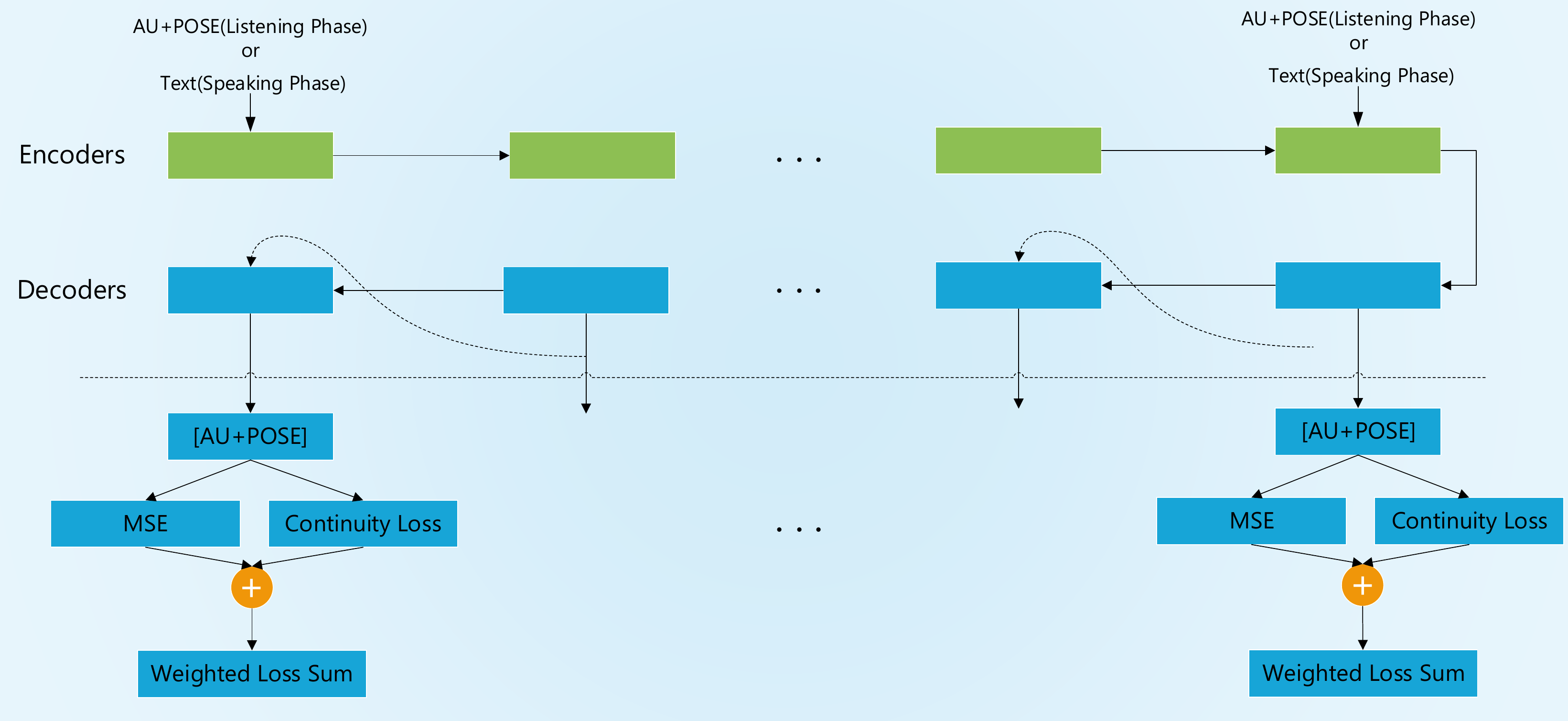}
	\end{center}
	\caption{Architecture of listening model and speaking model. In the listening phase, the encoder takes the AU+POSE as input. And in speaking phase, the encoder takes the text as input. In both phases, decoder outputs AU+POSE. MSE and Continuity Losses are employed to supervise the training for both phases.}
	\label{fig:3}
\end{figure*}

The speaking model takes the response text of listener as input and outputs the AU+POSE as the accompaniment of speech. We also base our speaking model on the Seq2Seq \cite{25} architecture(as shown in Figure \ref{fig:3}).

\textbf{Speaking encoder:}
There exist some popular word embedding methods, \eg, Word2Vec \cite{48},  GloVe \cite{31}, ELMo \cite{32} and BERT \cite{30} to represent the text. Here, for simplicity, we employ a pre-trained Word2Vec model to embed each word into a 200-D vector. Our speaking encoder is defined as

\begin{equation}\label{eqn:3}
\begin{aligned}
h_{text}^n = LSTM\left(\{ x_{text}^{n,l}\}_l\right)
\end{aligned}
\end{equation}
where the LSTM \cite{22} computes the forward sentence encodings, and applies a linear layer on top.

\textbf{Speaking decoder:}
The speaking decoder outputs an AU+POSE sequence used to synthesize the face sequence for the speaker (\ie, the previous listener).

\subsection{Seq2Seq training}

It is intuitive to use Mean Square Error (MSE) loss function to train our listening and speaking models, and the MSE loss can be formulated as

\begin{equation}\label{eqn:4}
\begin{aligned}
loss_{mse} = \frac{1}{n_{au}}\sum\limits_{i = 1}^{n_{au}} {\left\| au_{gt}^i-au_{pred}^i\right\|}_2 +\\
\frac{\gamma}{n_{pose}}\sum\limits_{i = 1}^{n_{pose}} {\left\| pose_{gt}^i-pose_{pred}^i\right\|}_2
\end{aligned}
\end{equation}
where $\gamma$ denotes the weight parameter, $au_{gt}^i$ and $au_{pred}^i$ denote the action-units vector in the ground truth and predicted sequence respectively, $pose_{gt}^i$ and $pose_{pred}^i$ denote the head-pose vector in the ground truth and predicted sequence. ${\left\|. \right\|}_2$ denotes the Euclidean norm and $n$ is the dimension of action-unit or head-pose.

However, the MSE loss cannot ensure the output continuity which is significant for generating realistic face sequences. To solve this problem, we introduce another loss: continuity loss to assist the Seq2Seq training. The continuity metric loss utilizes the constraints of adjacent frames which can considerably boost the performance of our models. The continuity metric loss can be formulated as:

\begin{equation}\label{eqn:5}
\begin{aligned}
\begin{array}{l}
loss(i,j)={\left\|\left(y_{gt}^{i-j}-y_{gt}^{i-j-1} \right) - \left(y_{pred}^{i-j} - y_{pred}^{i-j-1}\right)\right\|}_2 \\
loss_{con} = \frac{1}{n_{b}}\sum\limits_{i = n_{b}}^l max \left ( \left \{ loss(i,j)|j=1,\dots,n_{b}-1 \right \} \right)
\end{array}
\end{aligned}
\end{equation}
where $y_{gt}^{i}$ and $y_{pred}^{i}$ denote the AU+POSE vector in the ground truth and predicted sequence respectively, and $n_{b}$ denote the number of the adjacent frames used for calculating continuity metric loss. Thus, combining the MSE with continuity loss, we get the total loss as:

\begin{equation}\label{eqn:6}
\begin{aligned}
loss_{sum} = loss_{mse} + \alpha \cdot loss_{con}
\end{aligned}
\end{equation}
where $\alpha$ is a weight used to adjust the two loss components.

Note that the head-pose and each component of the action-unit have different variation ranges. Hence, both the input and output of the listening model are normalized during training, and the output of the speaking model is also normalized during training.

Moreover, all frames in the video clips involving face-to-face conversation are used for training our listening and speaking models to get a smoother and more realistic face sequences.

\begin{figure*}[htbp]
	\begin{center}
		\includegraphics[scale=0.18]{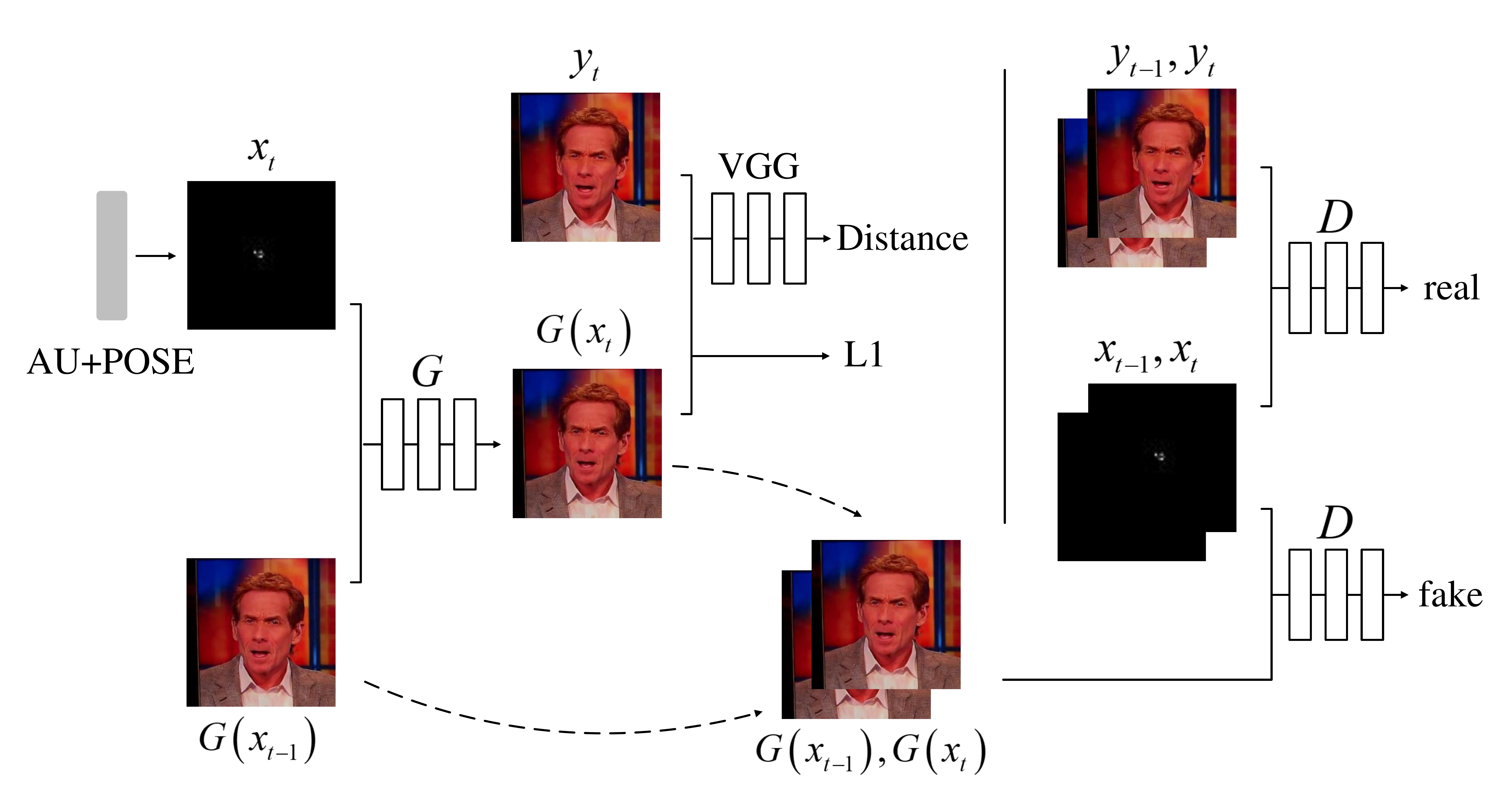}
	\end{center}
	\caption{Architecture of the face synthesizer. We first map AU+POSE to image $x$. In the training phase, we use pair $\left ( x_t,G \left (x_{t-1}\right )\right )$ to learn the mapping $G$ to generate $G\left (x_t \right )$. Discriminator $D$ then attempts to differentiate the ``real'' temporally coherent sequence $\left({{x_{t - 1}},{x_t},{y_{t - 1}},{y_t}} \right)$ from the ``fake'' sequence $\left( {{x_{t - 1}},{x_t},G\left( {{x_{t - 1}}} \right),G\left( {{x_t}} \right)} \right)$.}
	\label{fig:4}
\end{figure*}

\subsection{Face synthesizer}
With the AU+POSE as input, the face synthesizer aims to generate face images of the target person with corresponding postures and expressions. To accomplish this task, we adopt the state-of-the-art generative adversarial network Pix2PixHD \cite{16} to complete this transfer. Specifically, we employed a combined loss widely used in generation tasks:

\begin{equation}\label{eqn:7}
\begin{aligned}
L = L_{GAN}\left({G,D}\right)+L_{L1}\left({G\left(x\right),y}\right)+L_{VGG}\left({G\left(x\right),y}\right)
\end{aligned}
\end{equation}

where $L_{GAN}\left({G,D}\right)$ is the temporal smoothing adversarial loss presented in paper \cite{15}, which modified from Pix2PixHD \cite{16}. When synthesizing the current face image $G\left(x_t\right)$, Generator $G$ conditions on its corresponding current AU+POSE image $x$ and the previously synthesized frame $G\left(x_{t-1}\right)$ to obtain temporally smooth outputs. Discriminator $D$ then attempts differentiate the ``real'' temporally coherent sequence $\left({{x_{t - 1}},{x_t},{y_{t - 1}},{y_t}} \right)$ from the ``fake'' sequence $\left( {{x_{t - 1}},{x_t},G\left( {{x_{t - 1}}} \right),G\left( {{x_t}} \right)} \right)$:

\begin{equation}\label{eqn:8}
\begin{aligned}
{L_{GAN}}\left({G,D}\right) = {E}_{\left({x,y}\right)}\left[{\log D\left({{x_{t - 1}},{x_t},{y_{t - 1}},{y_t}} \right)} \right] +\\
{E}_x\left[ {\log \left( {1 - D\left( {{x_{t - 1}},{x_t},G\left( {{x_{t - 1}}} \right),G\left( {{x_t}} \right)} \right)} \right)} \right]
\end{aligned}
\end{equation}
where $L_{L1}$ is an $L1$ reconstruction loss that measures the pixel-level deviation between the synthesized image $G\left (x \right )$ and ground truth $y$. $L_{VGG}$ is the perceptual reconstruction loss which measures the $L2$ distance between relu2\_2 and relu3\_3 features of a pre-trained VGGNet \cite{18}.

The full transfer system  is shown in the Figure \ref{fig:4}. We use corresponding $\left({{x_{t - 1}},{x_t},{y_{t - 1}},{y_t}} \right)$ pairs to learn a mapping $G$ which synthesizes images of the target person given AU+POSE. It is noteworthy that we do not directly feed the AU+POSE feature values into the network. Instead, we fill the feature values into the center of an empty image $x$, where $x$ has the same size with the ground truth $y$. This allows our network to have spatial coordinate constraints when training, which can accelerate the convergence.

\section{Experimental results}\label{results}
\subsection{Implementation details}
We implement our seq2seq-based listening and speaking models by using TensorFlow framework \cite{29}. In listening phase, we construct a 4-layer encoder-decoder model using LSTM as shown in Figure \ref{fig:3}. We substantiate all these LSTMs with a 1000-d LSTM cell on top of a 1000-d embedding layer, followed by a linear layer with piecewise-linear relu6 to compute the final encoding. At each stage, both the decoder and the encoder have 10 cells cascaded, that is, corresponding consecutive 10 frames of data as input. The learning rate is set to 0.0001, the batch size is 16, and the ADAM \cite{24} is used as the optimizer. The training process stops after 80,000 iterations.

In speaking phase, the model uses only two layers of 1000-dimensional LSTM cell nested structure with relu6 on the activation function. The learning rate, batch size and optimizer selection are consistent with the listening phase, the $\gamma$ in Eq. \ref{eqn:4} is set to 8.1, the $\alpha$ in Eq. \ref{eqn:6} is set to 0.1, and the training stops after 40,000 iterations. At each stage, the encoder has 25 cells cascaded and the decoder has 370 cells cascaded. We first pre-train the encoder and decoder on a small subset of our training set, after that, the encoder is frozen and the decoder is trained. After these tasks are completed, we finally fine-tune the whole encoder-decoder network.

We train our face generation network for 100 epochs using the ADAM optimizer \cite{24} with lr=0.0002.

\subsection{Evaluation on Seq2Seq models}
We use MSE as well as cosine similarity in Eq. \ref{eqn:9} to evaluate the performance of the listening model and the speaking model separately on the test set. Correspondingly, test data are divided into two parts. In the listening test set, each sample is comprised of a sequence of AU+POSE of the speaker as input, and a sequence of AU+POSE of the listener as ground truth. In the speaking test set, the input is just a sequence of words spoken by the speaker, and the ground truth is a sequence of the speaker's AU+POSE. We compare our approach to the baseline: The classic Seq2Seq \cite{25} method that uses MSE loss only.

\begin{equation}\label{eqn:9}
\begin{aligned}
\begin{array}{l}
d_{mse} = \frac{1}{n}\sum {l_{mse}^i} \\
l_{mse} = \frac{1}{n_{au}}\sum\limits_{i = 1}^{n_{au}} {\left\| au_{gt}^i - au_{pred}^i\right\|}_2 + \\
\frac{1}{n_{pose}}\sum\limits_{i = 1}^{n_{pose}} {\left\| pose_{gt}^i - pose_{pred}^i \right\|}_2 \\
d_{con} = \frac{1}{n}\sum\limits_{i = 1}^n {\frac{y_{gt}^i \cdot y_{pred}^i} {\left\|y_{gt}^i \right\| \cdot \left\| y_{pred}^i \right\|}}
\end{array}
\end{aligned}
\end{equation}

\begin{table}[htbp]
	\small
	\begin{center}
		\renewcommand\tabcolsep{5mm}
		\begin{tabular}{|c|c|c|}
			\hline
			\multicolumn{1}{|c|}{\multirow{2}{*}{Method}} & \multicolumn{2}{c|}{Metric} \\
			\cline{2-3}  & MSE & Cosine\\
			\hline
			Baseline & 0.0565 & 0.983\\
			\hline
			Ours & 0.0540 & 0.992\\
			\hline
		\end{tabular}
	\end{center}
	\caption{Evaluation of training methods in listening phase.}
	\label{tab1}
\end{table}

\begin{table}[htbp]
	\small
	\begin{center}
		\renewcommand\tabcolsep{5mm}
		\begin{tabular}{|c|c|c|}
			\hline
			\multicolumn{1}{|c|}{\multirow{2}{*}{Method}} & \multicolumn{2}{c|}{Metric} \\
			\cline{2-3}  & MSE & Cosine\\
			\hline
			Baseline & 0.195 & 0.9635\\
			\hline
			Ours & 0.141 & 0.980\\
			\hline
		\end{tabular}
	\end{center}
	\caption{Evaluation of training methods in speaking phase.}
	\label{tab2}
\end{table}

In the listening phase as shown in Table \ref{tab1}, our method performs a little better than the baseline method in both MSE metric and cosine similarity metric. In the speaking phase as shown in Table \ref{tab2}, our method also outperforms the baseline method. It is noteworthy that the gap between out method and baseline in speaking phase is much larger than that in listening phase. One reason is that, when human is speaking, the diversity of facial action units and head poses is much more than when human is listening. So the variance of training data in speaking phase is much larger than that in listening phase. Consequently, the speaking model benefits more than the listening model from our proposed metric learning. The other reason is that, compared to the speaking model, the listening model is much deeper. Hence, the listening model has stronger learning ability which makes the gap between the baseline and our method in listening phase smaller than that in speaking phase.

\begin{figure*}[htbp]
	\begin{center}
		\includegraphics[width=0.9\linewidth]{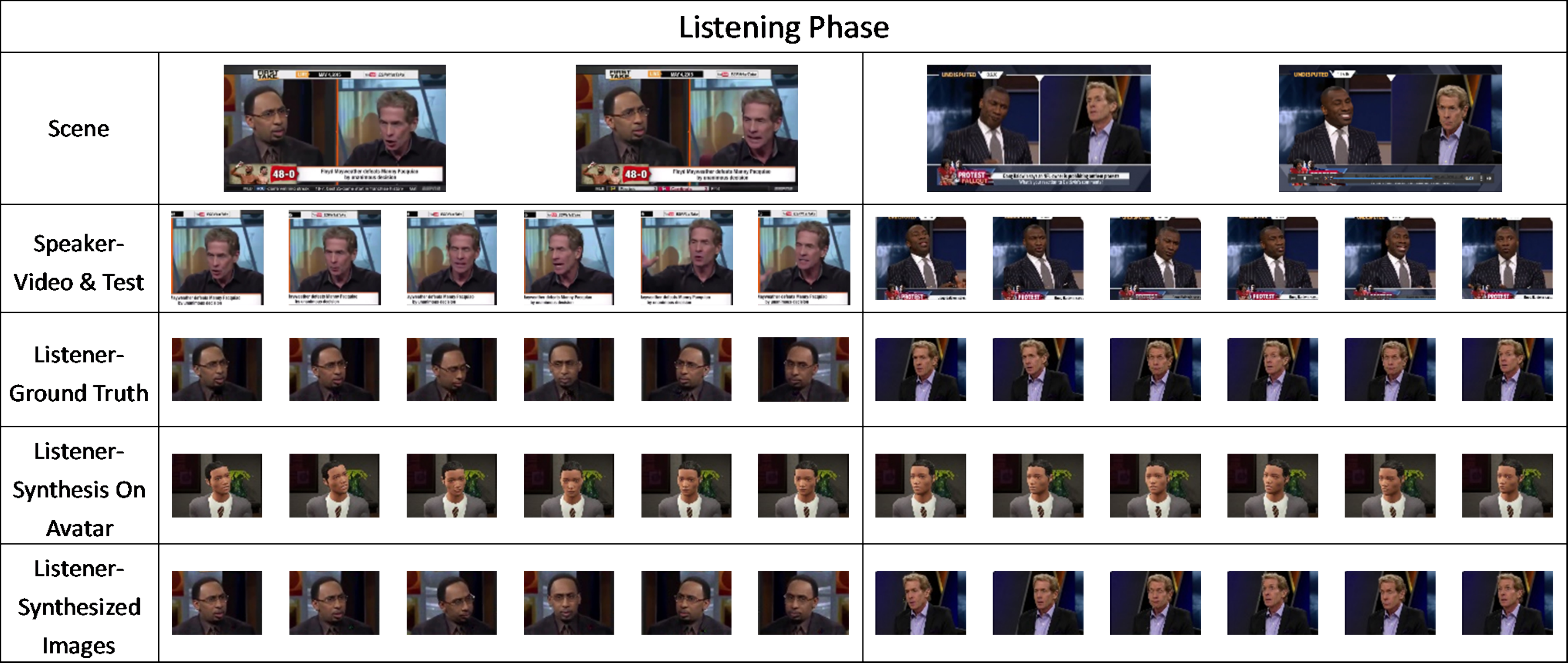}
	\end{center}
	\caption{Experimental results of our models in listening phase. Two scenes are shown here, each of which includes the conversation scene (the first row), the speaker's images (second row), the listener's ground truth images (third row), the listener's 3d avatar (fourth row), and the listener's synthesized images (last row).}
	\label{fig:5}
\end{figure*}

\begin{figure*}[htbp]
	\begin{center}
		\includegraphics[width=0.9\linewidth]{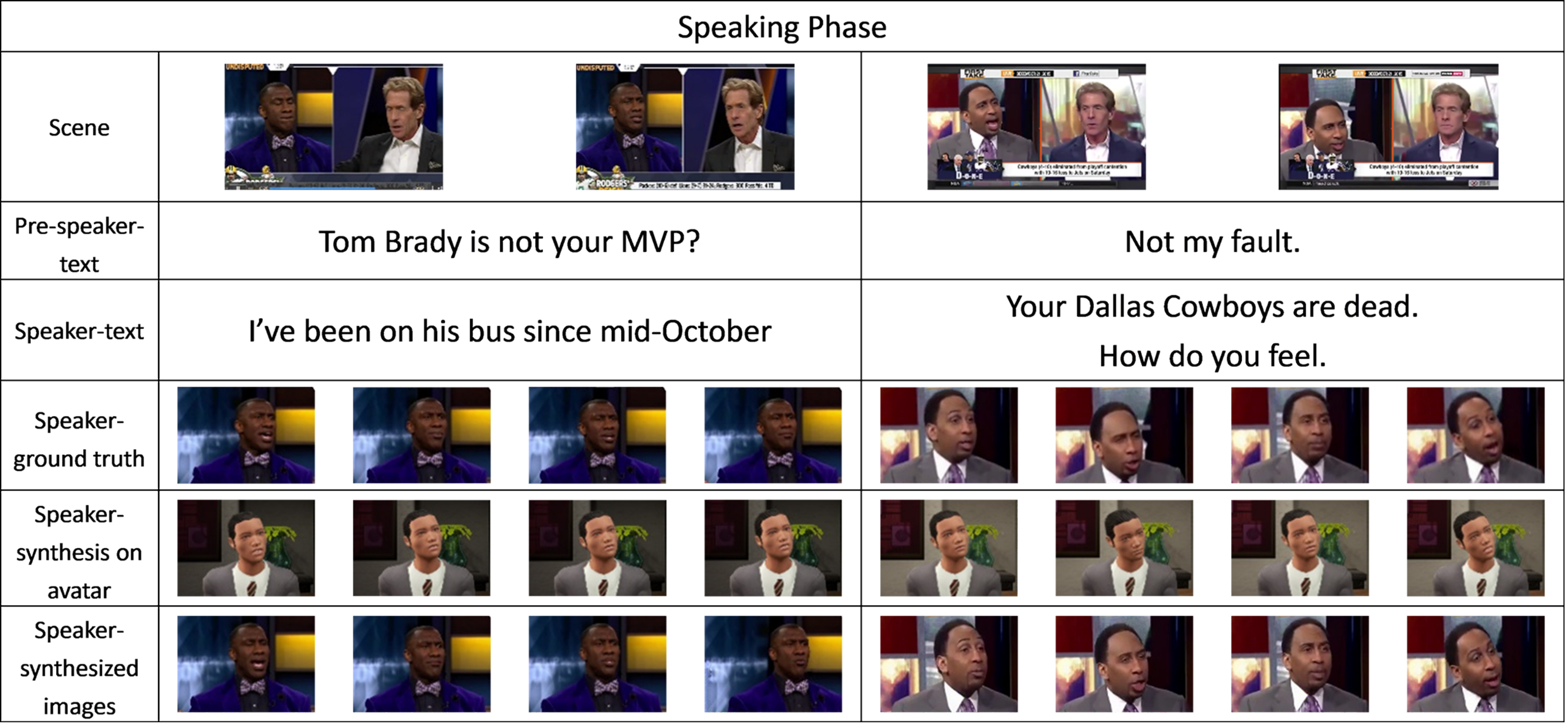}
	\end{center}
	\caption{Experimental results of our models in speaking phase. Two scenes are shown, each of which includes the conversation scene (the first row), the verbal cues of two persons in the scene (second row for the previous speaker and third row the present speaker), the speaker's ground truth images (fourth row), the speaker's 3d model based avatar (fifth row), and the speaker's synthesized images (last row).}
	\label{fig:6}
\end{figure*}

\subsection{Evaluation on avatar synthesizer}
We run facial action units and head pose extractor on the output of our avatar synthesizer, and compare these reconstructed action units and poses to the counterpart of the original input video. To assess the performance of our model, we employ the average error defined by

\begin{equation}\label{eqn:10}
\begin{aligned}
\begin{array}{l}
d_{au} = \frac{1}{m*{n_{au}}}\sum\limits_{j = 1}^m \sum\limits_{i = 1}^{n_{au}} {\left\| au_{gt}^{ij} - au_{reocnstructed}^{ij} \right\|}_1 \\
d_{pose} = \frac{1}{m*{n_{pose}}}\sum\limits_{j = 1}^m \sum\limits_{i = 1}^{n_{pose}} {\left\| pose_{gt}^{ij} - pose_{reocnstructed}^{ij}\right\|}_1
\end{array}
\end{aligned}
\end{equation}
where $au_{gt},au_{reocnstructed},pose_{gt},pose_{reocnstructed}$ denote the ground truth of action unit, the reconstructed action unit, ground truth of pose, the reconstructed pose and $n$ denotes the number of frames in test set.

\begin{table}[htbp]
	\small
	\begin{center}
		\renewcommand\tabcolsep{5mm}
		\begin{tabular}{|c|c|c|}
			\hline
			\multicolumn{1}{|c|}{\multirow{2}{*}{Phase}} & \multicolumn{2}{c|}{Metric} \\
			\cline{2-3}  & AU & Pose \\
			\hline
			Listening & 0.046 & 0.038 \\
			\hline
			Speaking & 0.105 & 0.073 \\
			\hline
		\end{tabular}
	\end{center}
	\caption{Evaluation of avatar synthesizer.}
	\label{tab3}
\end{table}

As shown in Table \ref{tab3}, in both the listening and speaking phases, the difference between the synthesized images and the original images is minor in terms of action unit and pose.

\subsection{Evaluation on End-to-End system}
Here we employ a 3D model based avatar directly driven by facial action unit and head pose as comparison. Four examples (two for listening and another two for speaking) are illustrated in Figure \ref{fig:5} and Figure \ref{fig:6} to show the comparison between ground truth and predictions of our models, as well as the results of 3D avatar and our synthesizer. As can be seen, the facial images generated by our synthesizer are more similar to the ground-truth, and more realistic than the 3D model based avatar.

\section{Conclusion}\label{summary}
 We present a realistic face-to-face conversation system based on deep neural networks. In listening phase, the speaker's face features including action unit and head pose are firstly extracted, then transformed into the avatar's face features by a seq2seq model, and finally used to generate realistic face images (avatar) by a pix2pix based GAN network. In speaking phase, the previous speaker's words are turned into word features by word2vec, then transformed into the avatar's face features by another seq2seq model, and finally used to generate realistic face images (avatar) by the pix2pix based GAN network. The model is trained and evaluated with the video data from the ESPN shows. For comparison, both the realistic avatar generated by the proposed models and the one driven by traditional 3D model based method are given. It shows that the proposed conversation system can generate natural and realistic avatars. In the future, multi-modal information will be considered to produce more personalized conversation avatars.

{\small
\bibliographystyle{ieee}
\bibliography{egbib}
}

\end{document}